\documentclass[conference]{IEEEtran}
\IEEEoverridecommandlockouts

\usepackage{cite}
\usepackage{amsmath,amssymb,amsfonts}
\usepackage{algorithmic}
\usepackage{graphicx}
\usepackage{textcomp}
\usepackage{xcolor}
\usepackage{booktabs}
\def\BibTeX{{\rm B\kern-.05em{\sc i\kern-.025em b}\kern-.08em
    T\kern-.1667em\lower.7ex\hbox{E}\kern-.125emX}}
\begin{document}

\title{Hybrid Quantum Temporal Convolutional Networks
\thanks{The views expressed in this article are those of the authors and do not represent the views of Wells Fargo. This article is for informational purposes only. Nothing contained in this article should be construed as investment advice. Wells Fargo makes no express or implied warranties and expressly disclaims all legal, tax, and accounting implications related to this article.}
}

\IEEEaftertitletext{\centering \vspace*{-2.0\baselineskip} \footnotesize \textsuperscript{*} These authors contributed equally \vspace{1.5\baselineskip}}

\author{\IEEEauthorblockN{Junghoon Justin Park*}
\IEEEauthorblockA{
\textit{Seoul National University}\\
Seoul, Korea \\
utopie9090@snu.ac.kr}
\and
\IEEEauthorblockN{Maria Pak*}
\IEEEauthorblockA{
\textit{Seoul National University}\\
Seoul, Korea \\
pakmasha99@snu.ac.kr}
\and
\IEEEauthorblockN{Sebin Lee*}
\IEEEauthorblockA{
\textit{Seoul National University}\\
Seoul, Korea \\
wallen@snu.ac.kr}
\and
\IEEEauthorblockN{Samuel Yen-Chi Chen}
\IEEEauthorblockA{\textit{Wells Fargo}\\
New York, USA \\
ycchen1989@ieee.org}
\and
\IEEEauthorblockN{Shinjae Yoo}
\IEEEauthorblockA{
\textit{Brookhaven National Laboratory}\\
New York, USA \\
sjyoo@bnl.gov}
\and
\IEEEauthorblockN{Huan-Hsin Tseng}
\IEEEauthorblockA{
\textit{Brookhaven National Laboratory}\\
New York, USA \\
htseng@bnl.gov}
\and
\IEEEauthorblockN{Jiook Cha}
\IEEEauthorblockA{
\textit{Seoul National University}\\
Seoul, Korea \\
connectome@snu.ac.kr}
}

\maketitle

\begin{abstract}
Quantum machine learning models for sequential data face scalability challenges with complex multivariate signals. We introduce the Hybrid Quantum Temporal Convolutional Network (HQTCN), which combines classical temporal windowing with a quantum convolutional neural network core. By applying a shared quantum circuit across temporal windows, HQTCN captures long-range dependencies while achieving significant parameter reduction. Evaluated on synthetic NARMA sequences and high-dimensional EEG time-series, HQTCN performs competitively with classical baselines on univariate data and outperforms all baselines on multivariate tasks. The model demonstrates particular strength under data-limited conditions, maintaining high performance with substantially fewer parameters than conventional approaches. These results establish HQTCN as a parameter-efficient approach for multivariate time-series analysis.
\end{abstract}


\section{Introduction}
Sequential data analysis presents significant challenges across many domains. While classical models have achieved considerable success, they often struggle to efficiently capture complex, long-range dependencies in high-dimensional time-series. Quantum machine learning (QML) offers a promising approach to address these limitations. Through quantum effects such as superposition and entanglement, QML models can process information in exponentially large Hilbert spaces \cite{Zhao}, potentially enabling more efficient learning of complex correlations in sequential data. Additionally, QML models have shown promise for data-efficient learning with relatively few parameters \cite{Caro2022}.

Recent QML research has explored quantum adaptations of classical sequential models. Quantum Long Short-Term Memory (QLSTM) \cite{QLSTM} and Quantum Recurrent Neural Networks (QRNN) \cite{LiQRNN, Bausch} demonstrated how variational quantum circuits (VQCs) could process temporal data. However, recurrent quantum architectures face inherent limitations as they rely on encoding the entire hidden state or input vector into a quantum circuit at each time step. While effective for univariate or low-dimensional data, this approach scales poorly to multivariate signals. For instance, encoding a 64-channel EEG signal into a quantum state requires either an unfeasible number of qubits for dense encoding or aggressive dimensionality reduction that discards critical information. Consequently, most QML time-series research remains restricted to low-dimensional synthetic benchmarks, failing to address complex multivariate tasks.

Furthermore, recurrent quantum architectures face severe implementation challenges on Noisy Intermediate-Scale Quantum (NISQ) devices. The sequential dependency of RNNs creates deep computational graphs that exacerbate noise accumulation, while Backpropagation-Through-Time on quantum hardware is computationally prohibitive \cite{QLSTM, Chen_QFWP}.

To bridge this gap, we propose the Hybrid Quantum Temporal Convolutional Network (HQTCN), which combines classical fully-connected layers with quantum convolutional neural network (QCNN) components. Instead of forcing the quantum circuit to manage temporal dynamics directly (as in QLSTMs), our approach employs a temporal sliding window strategy that applies dilated sampling to input sequences before processing them through shared quantum circuits. This temporal windowing mechanism enables the quantum components to effectively capture dependencies across different time scales while significantly reducing parameter overhead, making it suitable for scalable processing of high-dimensional multivariate time-series.

This architecture provides three distinct advantages:
\begin{enumerate}
    \item \textbf{Multivariate Scalability:} HQTCN is, to our knowledge, one of the first QML architectures capable of processing high-dimensional multivariate time-series (e.g., 64-channel EEG), a domain where QLSTMs are currently inapplicable.
    \item \textbf{Parameter Efficiency:} By sharing a compact quantum circuit across all temporal windows, HQTCN achieves competitive performance with orders of magnitude fewer parameters than classical baselines (e.g., $35\times$ fewer parameters than classical TCN).
    \item \textbf{Sample Efficiency:} Leveraging the high-dimensional feature space of the quantum kernel, HQTCN demonstrates superior generalization in data-limited regimes, significantly outperforming classical counterparts when training data is scarce ($N=10$ subjects).
\end{enumerate}

\section{Related Works}

\subsection{Recurrent Quantum Models}
Early quantum approaches to sequential data processing adapted classical recurrent architectures, leading to the development of QRNNs and QLSTMs \cite{QLSTM, LiQRNN, Bausch}. These models typically utilize VQCs as recurrent cells to process sequence elements iteratively, showing promise in simulation studies for forecasting tasks \cite{Padha, Khan}.

However, existing recurrent quantum models face both architectural and fundamental limitations. Architecturally, they are primarily designed for univariate or low-dimensional inputs. At each time step $t$, the model must encode the input vector $x_t \in \mathbb{R}^N$ into a quantum state. For high-dimensional multivariate signals—such as 64-channel EEG or dense financial portfolios—this encoding becomes prohibitively expensive. Standard angle embedding requires a number of qubits scaling linearly with the number of features $N$, which quickly exceeds the capacity of NISQ devices. Alternatively, amplitude embedding allows for logarithmic qubit scaling ($\mathcal{O}(\log N)$) but requires deep, complex state-preparation circuits that are difficult to train and highly susceptible to noise \cite{Sun2023}.

More critically, quantum mechanical constraints pose implementation challenges: the no-cloning theorem prevents perfect copying of quantum states, while accessing information from quantum hidden states requires measurements that collapse superpositions and destroy quantum information \cite{Kobayashi}. 

Practical implementations must rely on mid-circuit measurements and classical feed-forward operations, which introduce noise and computational overhead on current NISQ devices \cite{Viqueira}. These limitations motivate exploring novel quantum architectures that can better leverage quantum advantages for time-series analysis.

Consequently, standard QLSTMs and QRNNs effectively cannot scale to high-dimensional multivariate signals. This is not a minor architectural limitation; it excludes these QML models from the majority of real-world time-series applications, including biomedical signal processing (e.g., EEG, fMRI), multi-sensor industrial monitoring, and high-frequency trading. Furthermore, the sequential nature of these models necessitates Backpropagation-Through-Time, which incurs significant resource overheads on quantum hardware and complicates gradient calculation due to the need for repeated measurements \cite{QLSTM, Chen_QFWP}.

\subsection{Classical Temporal Convolutional Networks (TCNs)}
TCNs have emerged as an effective alternative to RNNs for sequence modeling \cite{TCN}. TCNs employ causal convolutions that respect temporal ordering and dilated convolutions that expand the receptive field by sampling inputs with increasing stride patterns. By stacking residual blocks with exponentially increasing dilation factors, TCNs efficiently capture long-range dependencies while enabling parallel processing during training.

Despite these advantages, classical TCNs may require substantial depth and parameters to model complex, high-dimensional time-series with intricate correlations. This computational overhead suggests potential benefits from quantum enhancement, where quantum superposition and entanglement could provide more efficient representations of complex temporal patterns.

\subsection{Quantum Convolutional Neural Network (QCNN)} \label{QCNN}
QCNNs extend classical CNNs to the quantum domain, offering advantages in parameter efficiency and noise resilience~\cite{Cong}. QCNNs employ a hierarchical structure alternating between convolutional and pooling layers implemented as quantum operations (Fig. \ref{fig_model}(a)).

The QCNN processes input data $x$ by encoding it into an $n$-qubit quantum state $\boldsymbol{\rho}(x)$, then applying $L$ layers of quantum operations. Each convolutional layer applies parameterized unitary transformations $\mathcal{U}^{(\ell)}(\boldsymbol{\theta})$ to overlapping qubits: 
\begin{equation}\label{E: parameterized ransformations}
    \boldsymbol{\rho} \mapsto \mathcal{U}^{(\ell)\dagger}(\boldsymbol{\theta}) \, \boldsymbol{\rho} \, \mathcal{U}^{(\ell)}(\boldsymbol{\theta})
\end{equation}

To mimic a classical pooling layer, which reduces input dimensions, a partial trace operation $\mathcal{P}_B$ is applied to trace out a subset of qubits $B$. Specifically, let $A = \{ 1, \ldots, n\}$ be the qubits (indices) of the entire system and a sub-system $B \subseteq A$, the trace operator $\mathcal{P}_B$ acting on density state $\boldsymbol{\rho}$ in $\mathcal{H}_A$ is defined via:
\begin{equation}\label{Def: trace}
 \langle v | \, \mathcal{P}_B(\boldsymbol{\rho}) \, | w \rangle_B := \sum_j \langle v \otimes e_j | \, \boldsymbol{\rho} \, | w \otimes e_j \rangle_A
\end{equation}
where $v, w $ are arbitrary quantum states in Hilbert space $\mathcal{H}_{A \setminus B}$ and $e_i$ is any basis in $\mathcal{H}_B$. Together, the actions of $\mathcal{P}^{(\ell)} $ and $ \mathcal{U}^{(\ell)}$ are combined as:
\begin{multline}\label{Def: joint action}
 \Bigl< v | \, \mathcal{P}_B\left( \mathcal{U}^{(\ell)\dagger}(\boldsymbol{\theta}) \, \boldsymbol{\rho} \, \mathcal{U}^{(\ell)}(\boldsymbol{\theta}) \right)  \, | w \Bigr>_B := \\
 \sum_j  \bigl< v \otimes e_j | \, \mathcal{U}^{(\ell)\dagger}(\boldsymbol{\theta}) \, \boldsymbol{\rho} \, \mathcal{U}^{(\ell)}(\boldsymbol{\theta}) \, | w \otimes e_j \bigr>_A
\end{multline}
The joint action of $\mathcal{P}^{(\ell)}$ and $\mathcal{U}^{(\ell)}$ is defined as an operator $\mathcal{C}^{(\ell)}$ via Eq.~(\ref{Def: joint action}).

Starting with an initial state $\boldsymbol{\rho}^{(0)}$, the state is transformed layer by layer:
\begin{equation}
\boldsymbol{\rho}^{(\ell)}=\mathcal{C}^{(\ell)}\big(\boldsymbol{\rho}^{(\ell - 1)}\big) \quad \text{for} \,\, \ell = 1, 2, ..., L
\end{equation}
The final output of the network is the expectation value of a Hermitian observable $H$ measured on the final state $\boldsymbol{\rho}^{(L)}_{\boldsymbol{\theta}}$:
\begin{equation}
f_{\boldsymbol{\theta}}(x) = \text{Tr}\big[H\boldsymbol{\rho}^{(L)}_{\boldsymbol{\theta}}(x)\big].
\label{QCNN_Equation}
\end{equation}
where the function bears the parameters ${\boldsymbol{\theta}}$ due to the parameterized layers involved in Eq.~(\ref{E: parameterized ransformations}).

QCNNs offer two key advantages for near-term quantum devices: parameter efficiency with only $\mathcal{O}(\log(n))$ variational parameters for $n$-qubits~\cite{Cong}, and provable barren plateau resilience that ensures trainability as the system size increases~\cite{Pesah}. These properties make QCNN a strong candidate for building a quantum analogue of the TCN.

\section{Hybrid Quantum Temporal Convolutional Networks}

\begin{figure*}[thb]
\begin{minipage}[b]{1.0\linewidth}
  \centering
  \centerline{\includegraphics[width=18cm]{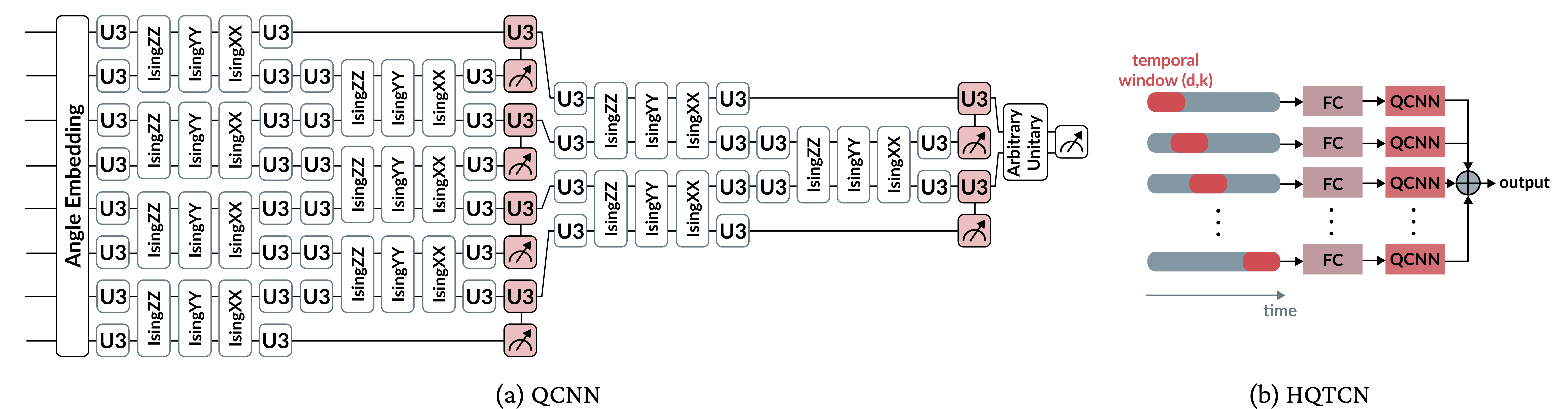}}
\end{minipage}
\caption{\textbf{Model Architectures.} (a) A standalone QCNN with hierarchical quantum convolutional and pooling layers. (b) The proposed HQTCN, which applies a shared QCNN circuit over a temporal sliding window before aggregating the outputs.}
\label{fig_model}
\end{figure*}

The HQTCN architecture overcomes the scalability limitations of recurrent quantum models by integrating classical temporal windowing with a shared QCNN core. As illustrated in Fig. \ref{fig_model}(b), the model processes multivariate time-series through four distinct stages: (1) Dilated Temporal Windowing, (2) Linear Embedding, (3) Quantum Feature Extraction, and (4) Output Aggregation.

\subsection{Dilated Temporal Windowing}
Let $X \in \mathbb{R}^{C \times T}$ denote a multivariate time series with $C$ channels and $T$ time steps. To capture long-range dependencies without the computational bottleneck of sequential recurrence, we apply a sliding window strategy inspired by classical TCN \cite{TCN}.

At each step $t$, we extract a local receptive field $W_t \in \mathbb{R}^{C \times K}$ by sampling $K$ points with a dilation factor $d \in \mathbb{N}$:
\begin{equation}
    \text{Indices}(t) = \{t - d(K-1), \dots, t - d, t\}
\end{equation}
The effective receptive field of a single window is $R = d(K-1) + 1$ time steps. This dilation allows the model to capture expanded temporal contexts without increasing the dimensionality of the quantum input. Crucially, unlike RNNs, these windows can be processed in parallel, offering significant speedups during training and inference.

\subsection{The Linear Embedding Interface}
A critical design choice in HQTCN is the interface between the classical high-dimensional input and the quantum circuit. Each flattened window $\mathbf{w}_t \in \mathbb{R}^{CK}$ is mapped to a quantum-compatible dimension $n$ (number of qubits) via a fully connected layer:
\begin{equation}
    \mathbf{e}_t = \mathbf{w}_t W_{fc} + \mathbf{b}_{fc},
\end{equation}
where $W_{fc} \in \mathbb{R}^{CK \times n}$ and $b_{fc} \in \mathbb{R}^{n}$ denote the weight matrix and bias of the classical fully-connected layer, respectively.

\textbf{Remark (Linearity)} It is worth noting that this embedding layer is a \textit{strictly linear projection} with no activation functions (e.g., ReLU, Tanh) or recurrent dynamics. This design ensures that the embedding layer performs only dimensionality reduction and cannot learn complex non-linear temporal correlations on its own. Consequently, any non-linear feature extraction and pattern recognition capability exhibited by the model must stem exclusively from the subsequent quantum circuit.

\subsection{Shared Quantum Convolutional Core}
The core computational unit is a shared QCNN circuit, which acts as a non-linear kernel. The embedding vector $\mathbf{e}_t$ is encoded into the quantum state via angle embedding, mapping the classical data into the high-dimensional Hilbert space of $n$ qubits.

The circuit $f_{\boldsymbol{\theta}}$ of Eq.~(\ref{QCNN_Equation}) processes the state using a sequence of quantum convolutional and pooling layers with trainable parameters $\boldsymbol{\theta} = \{\boldsymbol{\theta}_{conv}, \boldsymbol{\theta}_{pool}\}$. The final scalar output for the window is the expectation value of a measurement observable:
\begin{equation}
    o_t = f_{\boldsymbol{\theta}}(\mathbf{e}_t).
\end{equation}
\textbf{Remark (Weight Sharing)} To enforce temporal invariance and parameter efficiency, the \textit{same} quantum circuit parameters $\boldsymbol{\theta}$ are shared across all temporal windows $t$. This reduces the number of quantum parameters from $\mathcal{O}(T)$ (in a distinct-circuit approach) to $\mathcal{O}(1)$, independent of the sequence length. 

\subsection{Output Aggregation}
Processing the quantum circuit from $t = d(K-1)$ to $T-1$ yields the sequence of local quantum features $o_t = \{o_{d(K-1)}, \ldots, o_{T-1} \}$. A single prediction $\hat{Y}$ for the entire input is then obtained by aggregating these outputs, for example by averaging:
\begin{equation}
    \hat{Y} = \frac{1}{T - d(K-1)} \sum_{t=d(K-1)}^{T-1} o_t.
\end{equation}
This structure captures temporal dependencies by consistently applying the same quantum processing cell across the sequence.

\section{Experiment}

\subsection{Data}

\subsubsection{NARMA}
We evaluate time-series prediction using the Non-linear Autoregressive Moving Average (NARMA) dataset~\cite{NARMA}, a standard benchmark for sequential models~\cite{Atiya, Schrauwen}. The 10th-order NARMA sequence is generated by:
\begin{equation}
    y_t = 0.3y_{t-1}+0.05y_{t-1}\sum_{i=0}^9 y_{t-i-1} + 1.5u_{t-10}u_{t-1}+0.1
\end{equation}
where $u$ is an i.i.d. random input sequence. This task tests models' ability to capture complex long-term dependencies. We generated 240 time points, split into 70\% training, 15\% validation, and 15\% testing.

\subsubsection{PhysioNet EEG}
We use the motor-imagery subset from PhysioNet EEG~\cite{PhysioNet} recorded with BCI2000~\cite{PhysioNetEEG}. From 109 subjects, we selected left- versus right-hand imagery data from 50 randomly chosen subjects. The data was sampled at 80 Hz across 64 scalp channels for 249 time steps for the binary hand movement classification task.

\subsection{Experimental Setup}
We compare HQTCN against classical TCN~\cite{TCN}, LSTM~\cite{LSTM}, Transformer~\cite{Vaswani}, and QLSTM~\cite{QLSTM}. QLSTM was excluded
from PhysioNet experiments due to its limitations with multivariate data. We include a standalone QCNN~\cite{Cong} for ablation analysis of our temporal window approach.

Models were implemented using PennyLane~\cite{bergholm2022pennylane} and PyTorch on a Linux server with 128 CPU cores (256 threads), 503 GB of RAM, and an NVIDIA A100-PCIE GPU with 40 GB of memory. The software stack consisted of Python 3.11.7, PyTorch 2.5.0+cu121, and CUDA 12.1. All QCNN-based models used 8-qubits with two layers (convolution + pooling each).

Training hyperparameters were: NARMA (kernel size 5, dilation 2, learning rate 0.005, weight decay 0.0001); PhysioNet EEG (kernel size 12, dilation 3, learning rate 0.001, weight decay 0.0001).

\section{Results}

\begin{figure}[htb]
\begin{minipage}[b]{1.0\linewidth}
  \centering
  \centerline{\includegraphics[width=8.5cm]{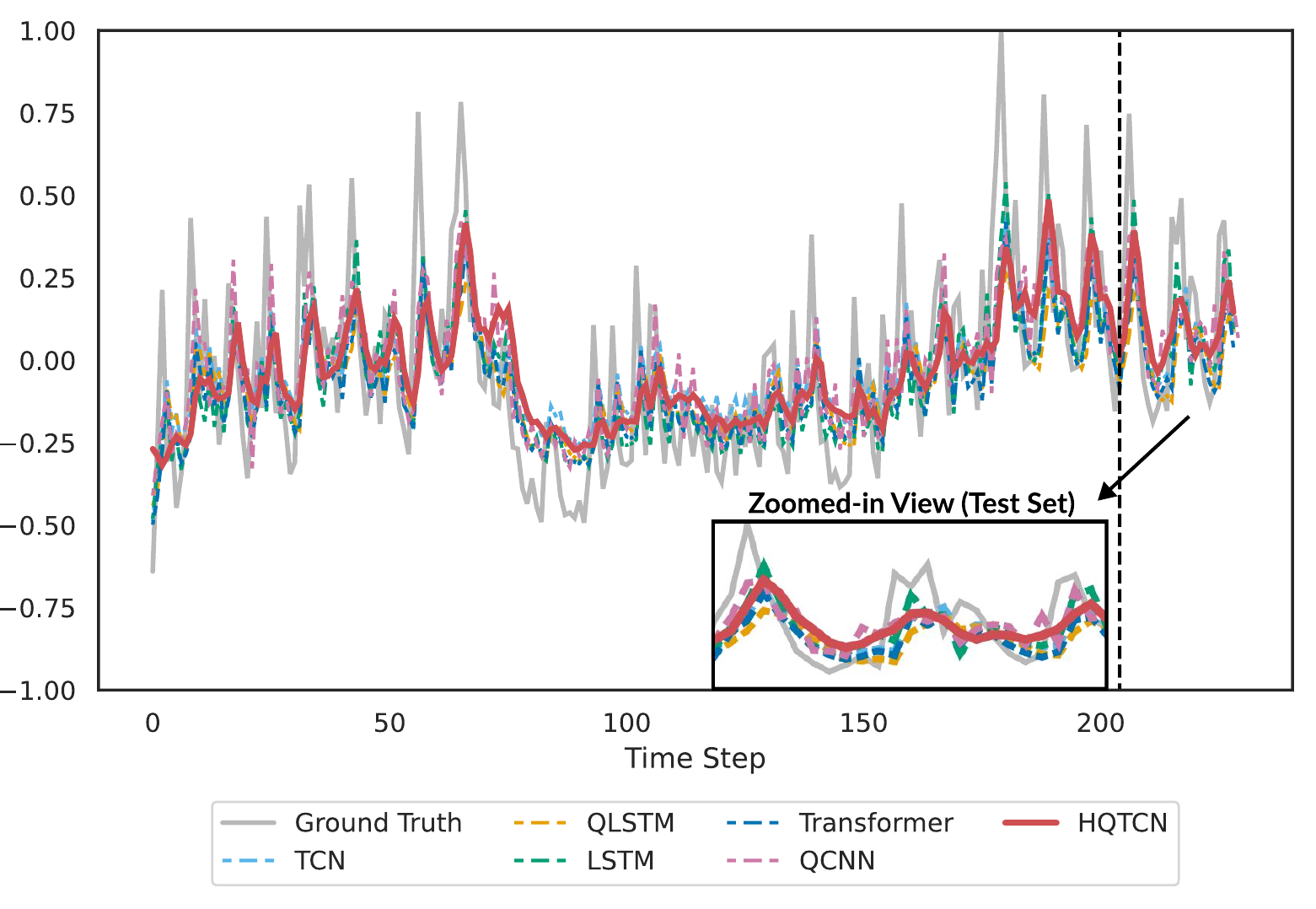}}
\end{minipage}
\caption{\textbf{NARMA Time-Series Prediction Results.} Comparison of HQTCN's predictions (red) against the ground truth (gray) and other models on the NARMA test set. The zoomed-in view highlights the test-set performance.}
\label{Fig_NARMA}
 \vspace{-0.5cm}
\end{figure}

\begin{table}[htp!]
    \centering
    \caption{
    \textbf{NARMA Dataset Results.} Comparison of mean test loss ($\pm$ std. dev. over three seeds) and the number of trainable parameters for all models on the NARMA prediction task.}
    \begin{tabular}{ccc}
    \toprule
        ~ & Test Loss  & \# of Parameters \\
        \midrule
        LSTM  & 0.0387 ± 0.0111  & 531  \\
        Transformer  & 0.0423 ± 0.0127  & 8,641  \\
        TCN  & 0.0380 ± 0.0032  & 10,785  \\
        QLSTM  & 0.0547 ± 0.0192  & 126 \\
        QCNN  & 0.0456 ± 0.0181  & 2,192  \\ 
        HQTCN & 0.0487 ± 0.0193 & 312 \\
    \bottomrule
    \end{tabular}
    \label{Table_NARMA}
\end{table}

\begin{table*}[!ht]
    \centering
    \caption{
    \textbf{PhysioNet EEG Analysis Results.} Test AUROC ($\pm$ std. dev. over three seeds) and parameter counts for all models on the EEG classification task across performance-optimized, parameter-constrained, and sample-constrained (N=10) settings.}
    \begin{tabular}{lccccc}
    \toprule
        ~ & \multicolumn{2}{c}{Performance-optimized} & \multicolumn{2}{c}{Parameter-constrained} & \multicolumn{1}{c}{Sample-constrained (N = 10)} \\ 
        \cmidrule(r){2-3} \cmidrule(r){4-5} \cmidrule(l){6-6}
        ~ & Test AUROC & \# of Parameters & Test AUROC & \# of Parameters & Test AUROC \\ 
        \midrule
        LSTM & 0.7552 ± 0.0182 & 66,625 & 0.7657 ± 0.0178 & 5,265 & 0.6582 ± 0.0832 \\ 
        Transformer & 0.7672 ± 0.0269 & 19,201 & 0.7618 ± 0.0261 & 7,729 & 0.6651 ± 0.0645 \\ 
        TCN & 0.7815 ± 0.0296 & 49,985 & 0.7755 ± 0.0245 & 5,857 & 0.5875 ± 0.0528 \\ 
        QCNN & 0.6649 ± 0.0342 & 127,760 & 0.6649 ± 0.0342 & 127,760 & 0.5112 ± 0.0072 \\ 
       \textbf{HQTCN} & \textbf{0.7929 ± 0.0383} & \textbf{6,416} & \textbf{0.7929 ± 0.0383} & \textbf{6,416} & \textbf{0.6713 ± 0.0515} \\
    \bottomrule
    \end{tabular}
    \label{Table_EEG}
\end{table*}

\subsection{NARMA: Extreme Parameter Efficiency}
We first evaluate the models on the univariate NARMA benchmark to assess the trade-off between predictive accuracy and model complexity. As shown in Table \ref{Table_NARMA} and Fig. \ref{Fig_NARMA}, the classical TCN achieves the lowest test loss ($0.0380 \pm 0.0032$). However, this performance comes at a high computational cost: the TCN requires 10,785 trainable parameters to model the sequence effectively.

In stark contrast, HQTCN achieves a comparable test loss of $0.0487 \pm 0.0193$ using only \textbf{312 parameters}. This represents a \textbf{97\% reduction} in model size (approximately $35\times$ fewer parameters) compared to the classical baseline. The result demonstrates that HQTCN can capture complex non-linear temporal dependencies with extreme parameter efficiency. This makes the architecture particularly suitable for resource-constrained edge computing environments where memory footprint is a primary bottleneck.

\subsection{PhysioNet EEG: Multivariate Superiority}
On the high-dimensional PhysioNet EEG dataset (Table \ref{Table_EEG}), HQTCN demonstrates its primary advantage. In the \textit{Performance-optimized} setting, where all baseline models were trained with their optimal number of parameters to maximize performance, HQTCN achieves the highest test AUROC of $0.7929 \pm 0.0383$ despite having significantly fewer parameters (6,416 total: 6,152 classical + 264 quantum) than classical baselines.

This advantage is further validated in the parameter-constrained scenario, where classical baseline models' parameters were reduced to approximately match HQTCN's parameter count ($\approx 6.4$k). HQTCN maintained superior performance against these parameter-matched baselines, demonstrating exceptional parameter efficiency.

The model's efficiency is particularly evident in data-limited scenarios. With only 10 subjects, HQTCN achieves an AUROC of 0.6713 ± 0.0515, significantly outperforming classical TCN ($0.5875 \pm 0.0528$) and QCNN ($0.5112 \pm 0.0072$).

Our temporal sliding window approach addresses a key limitation of standalone QCNN: since QCNNs are not designed for multivariate temporal data, they require extensive classical preprocessing layers that scale with input dimensionality. Our windowing architecture circumvents this issue, enabling HQTCN to process multivariate sequences with $20\times$ fewer parameters than standalone QCNN (6,416 vs 127,760) while improving generalizability. This efficiency makes HQTCN particularly effective for high-dimensional, multivariate time-series such as EEG signals, where it significantly outperforms baselines.

\subsection{Ablation Studies}
To isolate the contributions of specific architectural components, we conducted comprehensive ablation studies on the PhysioNet EEG dataset. We analyzed three key hyperparameters: dilation factor $d$, quantum circuit depth $L$, and embedding dimension $n$. Results are summarized in Table \ref{Table_Ablation}.

\begin{table}[h]
    \centering
    \caption{
    \textbf{Ablation studies on PhysioNet EEG.} We report Test AUROC ($\pm$ std. dev. over three seeds) varying one hyperparameter while fixing others to baseline values ($d=3, L=2, n=8$). Receptive field $= d \times (K-1) + 1$, where $K=12$.}
    \begin{tabular}{lcc}
    \toprule
        Dilation ($d$) & Test AUROC & Receptive Field \\
        \cmidrule(r){1-3}
        $d=1$ & $0.7859 \pm 0.0509$ & 12 time-steps \\
        $d=2$ & $0.7882 \pm 0.0506$ & 23 time-steps \\
        $\mathbf{d=3}$ \textbf{(Baseline)} & $\mathbf{0.7929 \pm 0.0383}$ & \textbf{34 time-steps} \\
        $d=4$ & $0.7887 \pm 0.0479$ & 45 time-steps \\
        \midrule
        Depth ($L$) & Test AUROC & Quantum Parameters \\
        \cmidrule(r){1-3}
        $L=1$ & $0.7888 \pm 0.0490$ & 132 \\
        $\mathbf{L=2}$ \textbf{(Baseline)} & $\mathbf{0.7929 \pm 0.0383}$ & \textbf{264} \\
        $L=3$ & $0.7872 \pm 0.0469$ & 396 \\
        \midrule
        Embedding ($n$) & Test AUROC & Total Parameters\\ 
        \cmidrule(r){1-3}
        $n=4$ & $0.7832 \pm 0.0517$ & 3,380\\
        $\mathbf{n=8}$ \textbf{(Baseline)} &  $\mathbf{0.7929 \pm 0.0383}$ & \textbf{6,416}\\ 
        $n=16$ & $0.7687 \pm 0.0409$ & 12,704 \\ 
    \bottomrule
    \end{tabular}
    \label{Table_Ablation}
\end{table}

\textbf{Impact of Temporal Receptive Field:} Performance improves with dilation up to $d=3$, then slightly decreases at $d=4$. With a kernel size of $K=12$ and dilation $d=3$, the effective receptive field covers 34 time steps ($\approx 0.42$ seconds at 80 Hz). This aligns well with the neurophysiologically relevant time-scales for motor imagery EEG. This validates our design choice and demonstrates that HQTCN effectively leverages dilated sampling for long-range temporal dependencies.

\textbf{Circuit Depth and NISQ Feasibility:} We observe that increasing the quantum circuit depth beyond $L=2$ yields no performance gain ($0.7872 \pm 0.0469$ for $L=3$ vs $0.7929 \pm 0.0383$ for $L=2$). This indicates that the temporal windowing mechanism, rather than circuit depth, is the primary driver of performance. This finding is significant for NISQ implementation, as it confirms that shallow, noise-resilient circuits are sufficient for this architecture.

\textbf{Embedding Dimension and Overfitting:} The model achieves optimal performance with an embedding size of $n=8$ qubits. Smaller embeddings (n=4) reduce parameters by 47\% with only a marginal performance drop ($0.7832 \pm 0.0517$), suggesting highly efficient information compression. Conversely, doubling the dimension to $n=16$ degrades performance ($0.7687 \pm 0.0409$), likely due to overfitting on limited training data given the exponentially larger Hilbert space.

\textbf{Key Takeaways:} Our ablation analysis supports three critical architectural insights:
\begin{enumerate}
    \item \textbf{Shallow Depth Sufficiency:} Performance does not improve with circuit depth ($L>2$), indicating that the bottleneck is not depth of the VQC but rather the temporal receptive field.
    \item \textbf{Optimal Expressivity:} An 8-qubit embedding strikes the optimal balance between feature expressivity and generalization; larger Hilbert spaces ($n=16$) lead to overfitting.
    \item \textbf{Mechanism of Action:} The effectiveness of HQTCN stems primarily from the \textit{temporal windowing design} which enables long-range dependency capture, rather than simply scaling the quantum circuit size.
\end{enumerate}

\section{Conclusion}

We introduced HQTCN, a novel hybrid quantum-classical model for multivariate time-series analysis. While achieving comparable performance to baselines on univariate NARMA data, HQTCN significantly outperformed all classical and quantum models on the high-dimensional EEG dataset.

The model's architecture proves critical to its success. Ablation studies confirm that integrating classical temporal sliding windows with quantum convolutional components is essential for effective temporal sequence processing, with HQTCN substantially outperforming standalone QCNN models.

HQTCN achieves superior performance with exceptional parameter efficiency, using from 3-20$\times$ fewer parameters than baseline models while maintaining competitive accuracy. The model also demonstrates strong sample efficiency, outperforming all baseline models when trained on limited data (10 subjects).

We attribute this high efficiency to the inherent properties of quantum computing. By leveraging superposition and entanglement, the quantum components of our model can access a vast feature space, enhancing the model's expressibility and allowing it to learn key patterns and generalize effectively from a limited number of parameters and samples \cite{Zhao, Caro2022, Park2025}.

These results suggest that HQTCN represents a promising approach toward practical quantum advantage in complex time-series analysis, particularly for high-dimensional multivariate data where parameter efficiency and sample efficiency are crucial.

\section*{Acknowledgment}

Special thanks to the members of the SNU Connectome Lab for their invaluable support to the development of HQTCN model. This work was supported by the National Research Foundation of Korea (NRF) grant funded by the Korea government (MSIT) (No. 2021R1C1C1006503, RS-2023-00266787, RS-2023-00265406, RS-2024-00421268, RS-2024-00342301, RS-2024-00435727, NRF-2021M3E5D2A01022515, and NRF-2021S1A3A2A02090597), by the Creative-Pioneering Researchers Program through Seoul National University (No. 200-20240057, 200-20240135). Additional support was provided by the Institute of Information \& Communications Technology Planning \& Evaluation (IITP) grant funded by the Korea government (MSIT) [No. RS-2021-II211343, 2021-0-01343, Artificial Intelligence Graduate School Program, Seoul National University] and by the Global Research Support Program in the Digital Field (RS-2024-00421268). This work was also supported by the Artificial Intelligence Industrial Convergence Cluster Development Project funded by the Ministry of Science and ICT and Gwangju Metropolitan City, by the Korea Brain Research Institute (KBRI) basic research program (25-BR-05-01), by the Korea Health Industry Development Institute (KHIDI) and the Ministry of Health and Welfare, Republic of Korea (HR22C1605), and by the Korea Basic Science Institute (National Research Facilities and Equipment Center) grant funded by the Ministry of Education (RS-2024-00435727). We acknowledge the National Supercomputing Center for providing supercomputing resources and technical support (KSC-2023-CRE-0568). An award for computer time was provided by the U.S. Department of Energy’s (DOE) ASCR Leadership Computing Challenge (ALCC). This research used resources of the National Energy Research Scientific Computing Center (NERSC), a DOE Office of Science User Facility, under ALCC award m4750-2024, and supporting resources at the Argonne and Oak Ridge Leadership Computing Facilities, U.S. DOE Office of Science user facilities at Argonne National Laboratory and Oak Ridge National Laboratory.

\bibliographystyle{IEEEbib}
\bibliography{strings,refs}

\end{document}